# Machine Learning Operations (MLOps): Overview, Definition, and Architecture

Dominik Kreuzberger
KIT
Germany
dominik.kreuzberger@alumni.kit.edu

Niklas Kühl
KIT
Germany
kuehl@kit.edu

Sebastian Hirschl
IBM[†]
Germany
sebastian.hirschl@de.ibm.com

## ABSTRACT

The final goal of all industrial machine learning (ML) projects is to develop ML products and rapidly bring them into production. However, it is highly challenging to automate and operationalize ML products and thus many ML endeavors fail to deliver on their expectations. The paradigm of Machine Learning Operations (MLOps) addresses this issue. MLOps includes several aspects, such as best practices, sets of concepts, and development culture. However, MLOps is still a vague term and its consequences for researchers and professionals are ambiguous. To address this gap, we conduct mixed-method research, including a literature review, a tool review, and expert interviews. As a result of these investigations, we provide an aggregated overview of the necessary principles, components, and roles, as well as the associated architecture and workflows. Furthermore, we furnish a definition of MLOps and highlight open challenges in the field. Finally, this work provides guidance for ML researchers and practitioners who want to automate and operate their ML products with a designated set of technologies.

## 1 Introduction

Machine Learning (ML) has become an important technique to leverage the potential of data and allows businesses to be more innovative [1], efficient [13], and sustainable [22]. However, the success of many productive ML applications in real-world settings falls short of expectations [21]. A large number of ML projects fail—with many ML proofs of concept never progressing as far as production [30]. From a research perspective, this does not come as a surprise as the ML community has focused extensively on the building of ML models, but not on (a) building production-ready ML products and (b) providing the necessary coordination of the resulting, often complex ML system components and infrastructure, including the roles required to automate and operate an ML system in a real-world setting [35]. For instance, in many industrial applications, data scientists still manage ML workflows manually to a great extent, resulting in many issues during the operations of the respective ML solution [26].

To address these issues, the goal of this work is to examine how manual ML processes can be automated and operationalized so that more ML proofs of concept can be brought into production. In this work, we explore the emerging ML engineering practice "Machine Learning Operations"—MLOps for short—precisely addressing the issue of designing and maintaining productive ML. We take a holistic perspective to gain a common understanding of the involved components, principles, roles, and architectures. While existing research sheds some light on various specific aspects of MLOps, a holistic conceptualization, generalization, and clarification of ML systems design are still missing. Different perspectives and conceptions of the term "MLOps" might lead to misunderstandings and miscommunication, which, in turn, can lead to errors in the overall setup of the entire ML system. Thus, we ask the research question:

**RQ:** What is MLOps?

To answer that question, we conduct a mixed-method research endeavor to (a) identify important principles of MLOps, (b) carve out functional core components, (c) highlight the roles necessary to successfully implement MLOps, and (d) derive a general architecture for ML systems design. In combination, these insights result in a definition of MLOps, which contributes to a common understanding of the term and related concepts.

In so doing, we hope to positively impact academic and practical discussions by providing clear guidelines for professionals and researchers alike with precise responsibilities. These insights can assist in allowing more proofs of concept to make it into production by having fewer errors in the system's design and, finally, enabling more robust predictions in real-world environments.

The remainder of this work is structured as follows. We will first elaborate on the necessary foundations and related work in the field. Next, we will give an overview of the utilized methodology, consisting of a literature review, a tool review, and an interview study. We then present the insights derived from the application of the methodology and conceptualize these by providing a unifying definition. We conclude the paper with a short summary, limitations, and outlook.

---

[†] This paper does not represent an official IBM statement

## 2 Foundations of DevOps

In the past, different software process models and development methodologies surfaced in the field of software engineering. Prominent examples include waterfall [37] and the agile manifesto [5]. Those methodologies have similar aims, namely to deliver production-ready software products. A concept called "DevOps" emerged in the years 2008/2009 and aims to reduce issues in software development [9,31]. DevOps is more than a pure methodology and rather represents a paradigm addressing social and technical issues in organizations engaged in software development. It has the goal of eliminating the gap between development and operations and emphasizes collaboration, communication, and knowledge sharing. It ensures automation with continuous integration, continuous delivery, and continuous deployment (CI/CD), thus allowing for fast, frequent, and reliable releases. Moreover, it is designed to ensure continuous testing, quality assurance, continuous monitoring, logging, and feedback loops. Due to the commercialization of DevOps, many DevOps tools are emerging, which can be differentiated into six groups [23,28]: collaboration and knowledge sharing (e.g., Slack, Trello, GitLab wiki), source code management (e.g., GitHub, GitLab), build process (e.g., Maven), continuous integration (e.g., Jenkins, GitLab CI), deployment automation (e.g., Kubernetes, Docker), monitoring and logging (e.g., Prometheus, Logstash). Cloud environments are increasingly equipped with ready-to-use DevOps tooling that is designed for cloud use, facilitating the efficient generation of value [38]. With this novel shift towards DevOps, developers need to care about what they develop, as they need to operate it as well. As empirical results demonstrate, DevOps ensures better software quality [34]. People in the industry, as well as academics, have gained a wealth of experience in software engineering using DevOps. This experience is now being used to automate and operationalize ML.

## 3 Methodology

To derive insights from the academic knowledge base while also drawing upon the expertise of practitioners from the field, we apply a mixed-method approach, as depicted in Figure 1. As a first step, we conduct a structured literature review [20,43] to obtain an overview of relevant research. Furthermore, we review relevant tooling support in the field of MLOps to gain a better understanding of the technical components involved. Finally, we conduct semi-structured interviews [33,39] with experts from different domains. On that basis, we conceptualize the term "MLOps" and elaborate on our findings by synthesizing literature and interviews in the next chapter ("Results").

### 3.1 Literature Review

To ensure that our results are based on scientific knowledge, we conduct a systematic literature review according to the method of Webster and Watson [43] and Kitchenham et al. [20]. After an initial exploratory search, we define our search query as follows: *((("DevOps" OR "CICD" OR "Continuous Integration" OR "Continuous Delivery" OR "Continuous Deployment") AND "Machine Learning") OR "MLOps" OR "CD4ML")*. We query the scientific databases of Google Scholar, Web of Science, Science Direct, Scopus, and the Association for Information Systems eLibrary. It should be mentioned that the use of DevOps for ML, MLOps, and continuous practices in combination with ML is a relatively new field in academic literature. Thus, only a few peer-reviewed studies are available at the time of this research. Nevertheless, to gain experience in this area, the search included non-peer-reviewed literature as well. The search was performed in May 2021 and resulted in 1,864 retrieved articles. Of those, we screened 194 papers in detail. From that group, 27 articles were selected based on our inclusion and exclusion criteria (e.g., the term MLOps or DevOps and CI/CD in combination with ML was described in detail, the article was written in English, etc.). All 27 of these articles were peer-reviewed.

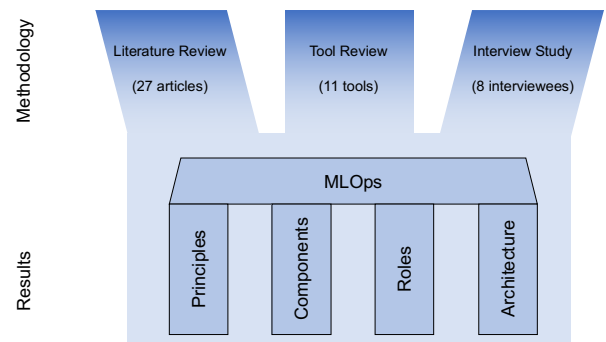

**Figure 1. Overview of the methodology**

### 3.2 Tool Review

After going through 27 articles and eight interviews, various open-source tools, frameworks, and commercial cloud ML services were identified. These tools, frameworks, and ML services were reviewed to gain an understanding of the technical components of which they consist. An overview of the identified tools is depicted in Table 1 of the Appendix.

### 3.3 Interview Study

To answer the research questions with insights from practice, we conduct semi-structured expert interviews according to Myers and Newman [33]. One major aspect in the research design of expert interviews is choosing an appropriate sample size [8]. We apply a theoretical sampling approach [12], which allows us to choose experienced interview partners to obtain high-quality data. Such data can provide meaningful insights with a limited number of interviews. To get an adequate sample group and reliable insights, we use LinkedIn—a social network for professionals—to identify experienced ML professionals with profound MLOps knowledge on a global level. To gain insights from various perspectives, we choose interview partners from different organizations and industries, different countries and nationalities, as well as different genders. Interviews are conducted until no new categories and concepts emerge in the analysis of the data. In total, we conduct eight interviews with experts (α - θ), whose details are depicted in Table 2 of the Appendix. According to Glaser and Strauss [5, p.61], this stage is called "theoretical saturation." All interviews are conducted between June and August 2021.

With regard to the interview design, we prepare a semi-structured guide with several questions, documented as an interview script [33]. During the interviews, "soft laddering" is used with "how" and "why" questions to probe the interviewees' means-end chain [39]. This methodical approach allowed us to gain additional insight into the experiences of the interviewees when required. All interviews are recorded and then transcribed. To evaluate the interview transcripts, we use an open coding scheme [8].

## 4 Results

We apply the described methodology and structure our resulting insights into a presentation of important principles, their resulting instantiation as components, the description of necessary roles, as well as a suggestion for the architecture and workflow resulting from the combination of these aspects. Finally, we derive the conceptualization of the term and provide a definition of MLOps.

### 4.1 Principles

A principle is viewed as a general or basic truth, a value, or a guide for behavior. In the context of MLOps, a principle is a guide to how things should be realized in MLOps and is closely related to the term "best practices" from the professional sector. Based on the outlined methodology, we identified nine principles required to realize MLOps. Figure 2 provides an illustration of these principles and links them to the components with which they are associated.

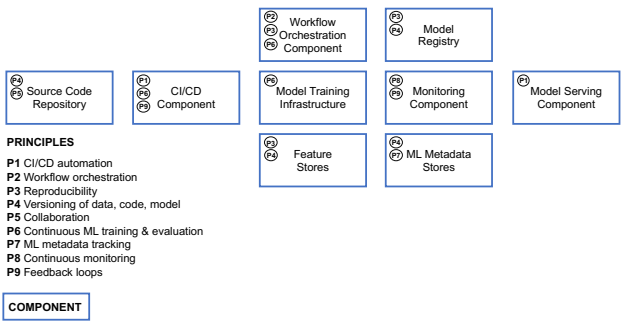

**Figure 2. Implementation of principles within technical components**

**P1 CI/CD automation.** CI/CD automation provides continuous integration, continuous delivery, and continuous deployment. It carries out the build, test, delivery, and deploy steps. It provides fast feedback to developers regarding the success or failure of certain steps, thus increasing the overall productivity [15,17,26,27,35,42,46] [α, β, θ].

**P2 Workflow orchestration.** Workflow orchestration coordinates the tasks of an ML workflow pipeline according to directed acyclic graphs (DAGs). DAGs define the task execution order by considering relationships and dependencies [14,17,26,32,40,41] [α, β, γ, δ, ζ, η].

**P3 Reproducibility.** Reproducibility is the ability to reproduce an ML experiment and obtain the exact same results [14,32,40,46] [α, β, δ, ε, η].

**P4 Versioning.** Versioning ensures the versioning of data, model, and code to enable not only reproducibility, but also traceability (for compliance and auditing reasons) [14,32,40,46] [α, β, δ, ε, η].

**P5 Collaboration.** Collaboration ensures the possibility to work collaboratively on data, model, and code. Besides the technical aspect, this principle emphasizes a collaborative and communicative work culture aiming to reduce domain silos between different roles [14,26,40] [α, δ, θ].

**P6 Continuous ML training & evaluation.** Continuous training means periodic retraining of the ML model based on new feature data. Continuous training is enabled through the support of a monitoring component, a feedback loop, and an automated ML workflow pipeline. Continuous training always includes an evaluation run to assess the change in model quality [10,17,19,46] [β, δ, η, θ].

**P7 ML metadata tracking/logging.** Metadata is tracked and logged for each orchestrated ML workflow task. Metadata tracking and logging is required for each training job iteration (e.g., training date and time, duration, etc.), including the model specific metadata—e.g., used parameters and the resulting performance metrics, model lineage: data and code used—to ensure the full traceability of experiment runs [26,27,29,32,35] [α, β, δ, ε, ζ, η, θ].

**P8 Continuous monitoring.** Continuous monitoring implies the periodic assessment of data, model, code, infrastructure resources, and model serving performance (e.g., prediction accuracy) to detect potential errors or changes that influence the product quality [4,7,10,27,29,42,46] [α, β, γ, δ, ε, ζ, η].

**P9 Feedback loops.** Multiple feedback loops are required to integrate insights from the quality assessment step into the development or engineering process (e.g., a feedback loop from the experimental model engineering stage to the previous feature engineering stage). Another feedback loop is required from the monitoring component (e.g., observing the model serving performance) to the scheduler to enable the retraining [4,6,7,17,27,46] [α, β, δ, ζ, η, θ].

### 4.2 Technical Components

After identifying the principles that need to be incorporated into MLOps, we now elaborate on the precise components and implement them in the ML systems design. In the following, the components are listed and described in a generic way with their essential functionalities. The references in brackets refer to the respective principles that the technical components are implementing.

**C1 CI/CD Component (P1, P6, P9).** The CI/CD component ensures continuous integration, continuous delivery, and continuous deployment. It takes care of the build, test, delivery, and deploy steps. It provides rapid feedback to developers regarding the success or failure of certain steps, thus increasing the overall productivity [10,15,17,26,35,46] [α, β, γ, ε, ζ, η]. Examples are Jenkins [17,26] and GitHub actions (η).

**C2 Source Code Repository (P4, P5).** The source code repository ensures code storing and versioning. It allows multiple developers to commit and merge their code [17,25,42,44,46] [α, β, γ, ζ, θ]. Examples include Bitbucket [11] [ζ], GitLab [11,17] [ζ], GitHub [25] [ζ ,η], and Gitea [46].

**C3 Workflow Orchestration Component (P2, P3, P6).** The workflow orchestration component offers task orchestration of an ML workflow via directed acyclic graphs (DAGs). These graphs represent execution order and artifact usage of single steps of the workflow [26,32,35,40,41,46] [α, β, γ, δ, ε, ζ, η]. Examples include Apache Airflow [α, ζ], Kubeflow Pipelines [ζ], Luigi [ζ], AWS SageMaker Pipelines [β], and Azure Pipelines [ε].

**C4 Feature Store System (P3, P4).** A feature store system ensures central storage of commonly used features. It has two databases configured: One database as an offline feature store to serve features with normal latency for experimentation, and one database as an online store to serve features with low latency for predictions in production [10,14] [α, β, ζ, ε, θ]. Examples include Google Feast [ζ], Amazon AWS Feature Store [β, ζ], Tecton.ai and Hopswork.ai [ζ]. This is where most of the data for training ML models will come from. Moreover, data can also come directly from any kind of data store.

**C5 Model Training Infrastructure (P6).** The model training infrastructure provides the foundational computation resources, e.g., CPUs, RAM, and GPUs. The provided infrastructure can be either distributed or non-distributed. In general, a scalable and distributed infrastructure is recommended [7,10,24–26,29,40,45,46] [δ, ζ, η, θ]. Examples include local machines (not scalable) or cloud computation [7] [η, θ], as well as non-distributed or distributed computation (several worker nodes) [25,27]. Frameworks supporting computation are Kubernetes [η, θ] and Red Hat OpenShift [γ].

**C6 Model Registry (P3, P4).** The model registry stores centrally the trained ML models together with their metadata. It has two main functionalities: storing the ML artifact and storing the ML metadata (see C7) [4,6,14,17,26,27] [α, β, γ, ε, ζ, θ]. Advanced storage examples include MLflow [α, η, ζ], AWS SageMaker Model Registry [ζ], Microsoft Azure ML Model Registry [ζ], and Neptune.ai [α]. Simple storage examples include Microsoft Azure Storage, Google Cloud Storage, and Amazon AWS S3 [17].

**C7 ML Metadata Stores (P4, P7).** ML metadata stores allow for the tracking of various kinds of metadata, e.g., for each orchestrated ML workflow pipeline task. Another metadata store can be configured within the model registry for tracking and logging the metadata of each training job (e.g., training date and time, duration, etc.), including the model specific metadata—e.g., used parameters and the resulting performance metrics, model lineage: data and code used [14,25–27,32] [α, β, δ, ζ, θ]. Examples include orchestrators with built-in metadata stores tracking each step of experiment pipelines [α] such as Kubeflow Pipelines [α,ζ], AWS SageMaker Pipelines [α,ζ], Azure ML, and IBM Watson Studio [γ]. MLflow provides an advanced metadata store in combination with the model registry [32,35].

**C8 Model Serving Component (P1).** The model serving component can be configured for different purposes. Examples are online inference for real-time predictions or batch inference for predictions using large volumes of input data. The serving can be provided, e.g., via a REST API. As a foundational infrastructure layer, a scalable and distributed model serving infrastructure is recommended [7,11,25,40,45,46] [α, β, δ, ζ, η, θ]. One example of a model serving component configuration is the use of Kubernetes and Docker technology to containerize the ML model, and leveraging a Python web application framework like Flask [17] with an API for serving [α]. Other Kubernetes supported frameworks are KServing of Kubeflow [α], TensorFlow Serving, and Seldion.io serving [40]. Inferencing could also be realized with Apache Spark for batch predictions [θ]. Examples of cloud services include Microsoft Azure ML REST API [ε], AWS SageMaker Endpoints [α, β], IBM Watson Studio [γ], and Google Vertex AI prediction service [δ].

**C9 Monitoring Component (P8, P9).** The monitoring component takes care of the continuous monitoring of the model serving performance (e.g., prediction accuracy). Additionally, monitoring of the ML infrastructure, CI/CD, and orchestration are required [7,10,17,26,29,36,46] [α, ζ, η, θ]. Examples include Prometheus with Grafana [η, ζ], ELK stack (Elasticsearch, Logstash, and Kibana) [α, η, ζ], and simply TensorBoard [θ]. Examples with built-in monitoring capabilities are Kubeflow [θ], MLflow [η], and AWS SageMaker model monitor or cloud watch [ζ].

### 4.3 Roles

After describing the principles and their resulting instantiation of components, we identify necessary roles in order to realize MLOps in the following. MLOps is an interdisciplinary group process, and the interplay of different roles is crucial to design, manage, automate, and operate an ML system in production. In the following, every role, its purpose, and related tasks are briefly described:

**R1 Business Stakeholder** (similar roles: Product Owner, Project Manager). The business stakeholder defines the business goal to be achieved with ML and takes care of the communication side of the business, e.g., presenting the return on investment (ROI) generated with an ML product [17,24,26] [α, β, δ, θ].

**R2 Solution Architect** (similar role: IT Architect). The solution architect designs the architecture and defines the technologies to be used, following a thorough evaluation [17,27] [α, ζ].

**R3 Data Scientist** (similar roles: ML Specialist, ML Developer). The data scientist translates the business problem into an ML problem and takes care of the model engineering, including the selection of the best-performing algorithm and hyperparameters [7,14,26,29] [α, β, γ, δ, ε, ζ, η, θ].

**R4 Data Engineer** (similar role: DataOps Engineer). The data engineer builds up and manages data and feature engineering pipelines. Moreover, this role ensures proper data ingestion to the databases of the feature store system [14,29,41] [α, β, γ, δ, ε, ζ, η, θ].

**R5 Software Engineer.** The software engineer applies software design patterns, widely accepted coding guidelines, and best practices to turn the raw ML problem into a well-engineered product [29] [α, γ].

**R6 DevOps Engineer.** The DevOps engineer bridges the gap between development and operations and ensures proper CI/CD automation, ML workflow orchestration, model deployment to production, and monitoring [14–16,26] [α, β, γ, ε, ζ, η, θ].

**R7 ML Engineer/MLOps Engineer.** The ML engineer or MLOps engineer combines aspects of several roles and thus has cross-domain knowledge. This role incorporates skills from data scientists, data engineers, software engineers, DevOps engineers, and backend engineers (see Figure 3). This cross-domain role builds up and operates the ML infrastructure, manages the automated ML workflow pipelines and model deployment to production, and monitors both the model and the ML infrastructure [14,17,26,29] [α, β, γ, δ, ε, ζ, η, θ].

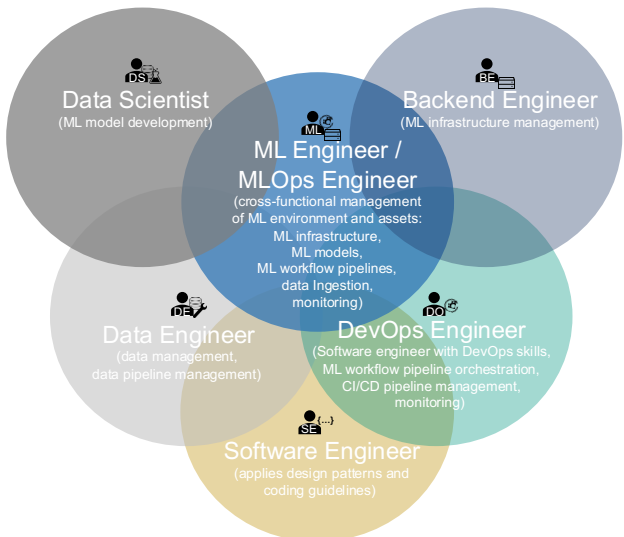

**Figure 3. Roles and their intersections contributing to the MLOps paradigm**

## 5 Architecture and Workflow

On the basis of the identified principles, components, and roles, we derive a generalized MLOps end-to-end architecture to give ML researchers and practitioners proper guidance. It is depicted in Figure 4. Additionally, we depict the workflows, i.e., the sequence in which the different tasks are executed in the different stages. The artifact was designed to be technology-agnostic. Therefore, ML researchers and practitioners can choose the best-fitting technologies and frameworks for their needs.

As depicted in Figure 4, we illustrate an end-to-end process, from MLOps project initiation to the model serving. It includes (A) the MLOps project initiation steps; (B) the feature engineering pipeline, including the data ingestion to the feature store; (C) the experimentation; and (D) the automated ML workflow pipeline up to the model serving.

**(A) MLOps project initiation.** (1) The business stakeholder (R1) analyzes the business and identifies a potential business problem that can be solved using ML. (2) The solution architect (R2) defines the architecture design for the overall ML system and, decides on the technologies to be used after a thorough evaluation. (3) The data scientist (R3) derives an ML problem—such as whether regression or classification should be used—from the business goal. (4) The data engineer (R4) and the data scientist (R3) work together in an effort to understand which data is required to solve the problem. (5) Once the answers are clarified, the data engineer (R4) and data scientist (R3) collaborate to locate the raw data sources for the initial data analysis. They check the distribution, and quality of the data, as well as performing validation checks. Furthermore, they ensure that the incoming data from the data sources is labeled, meaning that a target attribute is known, as this is a mandatory requirement for supervised ML. In this example, the data sources already had labeled data available as the labeling step was covered during an upstream process.

**(B1) Requirements for feature engineering pipeline.** The features are the relevant attributes required for model training. After the initial understanding of the raw data and the initial data analysis, the fundamental requirements for the feature engineering pipeline are defined, as follows: (6) The data engineer (R4) defines the data transformation rules (normalization, aggregations) and cleaning rules to bring the data into a usable format. (7) The data scientist (R3) and data engineer (R4) together define the feature engineering rules, such as the calculation of new and more advanced features based on other features. These initially defined rules must be iteratively adjusted by the data scientist (R3) either based on the feedback coming from the experimental model engineering stage or from the monitoring component observing the model performance.

**(B2) Feature engineering pipeline.** The initially defined requirements for the feature engineering pipeline are taken by the data engineer (R4) and software engineer (R5) as a starting point to build up the prototype of the feature engineering pipeline. The initially defined requirements and rules are updated according to the iterative feedback coming either from the experimental model engineering stage or from the monitoring component observing the model's performance in production. As a foundational requirement, the data engineer (R4) defines the code required for the CI/CD (C1) and orchestration component (C3) to ensure the task orchestration of the feature engineering pipeline. This role also defines the underlying infrastructure resource configuration. (8) First, the feature engineering pipeline connects to the raw data, which can be (for instance) streaming data, static batch data, or data from any cloud storage. (9) The data will be extracted from the data sources. (10) The data preprocessing begins with data transformation and cleaning tasks. The transformation rule artifact defined in the requirement gathering stage serves as input for this task, and the main aim of this task is to bring the data into a usable format. These transformation rules are continuously improved based on the feedback.

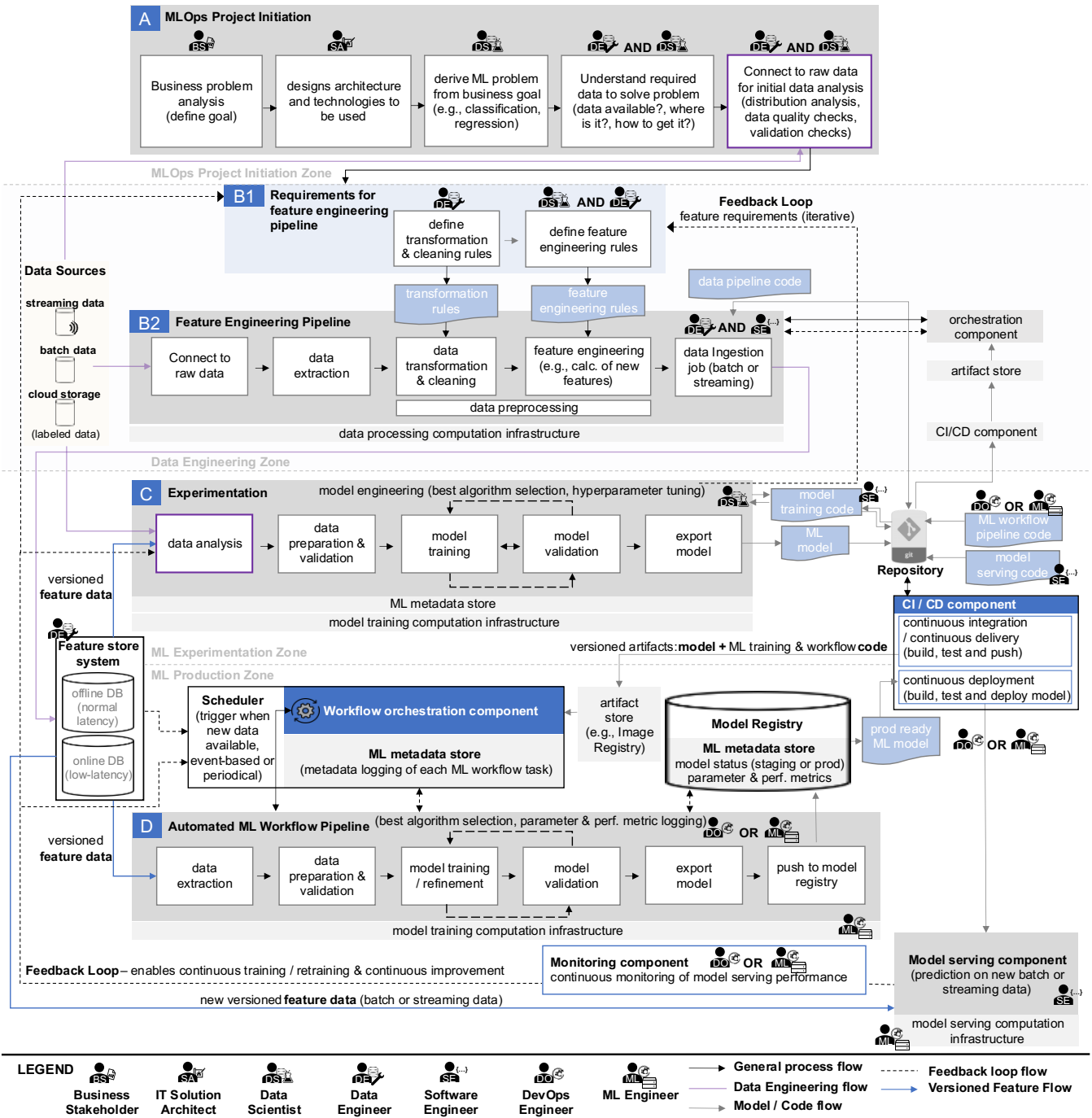

**Figure 4. End-to-end MLOps architecture and workflow with functional components and roles**

(11) The feature engineering task calculates new and more advanced features based on other features. The predefined feature engineering rules serve as input for this task. These feature engineering rules are continuously improved based on the feedback. (12) Lastly, a data ingestion job loads batch or streaming data into the feature store system (C4). The target can either be the offline or online database (or any kind of data store).

**(C) Experimentation.** Most tasks in the experimentation stage are led by the data scientist (R3). The data scientist is supported by the software engineer (R5). (13) The data scientist (R3) connects to the feature store system (C4) for the data analysis. (Alternatively, the data scientist (R3) can also connect to the raw data for an initial analysis.) In case of any required data adjustments, the data scientist (R3) reports the required changes back to the data engineering zone (feedback loop).

(14) Then the preparation and validation of the data coming from the feature store system is required. This task also includes the train and test split dataset creation. (15) The data scientist (R3) estimates the best-performing algorithm and hyperparameters, and the model training is then triggered with the training data (C5). The software engineer (R5) supports the data scientist (R3) in the creation of well-engineered model training code. (16) Different model parameters are tested and validated interactively during several rounds of model training. Once the performance metrics indicate good results, the iterative training stops. The best-performing model parameters are identified via parameter tuning. The model training task and model validation task are then iteratively repeated; together, these tasks can be called "model engineering." The model engineering aims to identify the best-performing algorithm and hyperparameters for the model. (17) The data scientist (R3) exports the model and commits the code to the repository.

As a foundational requirement, either the DevOps engineer (R6) or the ML engineer (R7) defines the code for the (C2) automated ML workflow pipeline and commits it to the repository. Once either the data scientist (R3) commits a new ML model or the DevOps engineer (R6) and the ML engineer (R7) commits new ML workflow pipeline code to the repository, the CI/CD component (C1) detects the updated code and triggers automatically the CI/CD pipeline carrying out the build, test, and delivery steps. The build step creates artifacts containing the ML model and tasks of the ML workflow pipeline. The test step validates the ML model and ML workflow pipeline code. The delivery step pushes the versioned artifact(s)—such as images—to the artifact store (e.g., image registry).

**(D) Automated ML workflow pipeline.** The DevOps engineer (R6) and the ML engineer (R7) take care of the management of the automated ML workflow pipeline. They also manage the underlying model training infrastructure in the form of hardware resources and frameworks supporting computation such as Kubernetes (C5). The workflow orchestration component (C3) orchestrates the tasks of the automated ML workflow pipeline. For each task, the required artifacts (e.g., images) are pulled from the artifact store (e.g., image registry). Each task can be executed via an isolated environment (e.g., containers). Finally, the workflow orchestration component (C3) gathers metadata for each task in the form of logs, completion time, and so on.

Once the automated ML workflow pipeline is triggered, each of the following tasks is managed automatically: (18) automated pulling of the versioned features from the feature store systems (data extraction). Depending on the use case, features are extracted from either the offline or online database (or any kind of data store). (19) Automated data preparation and validation; in addition, the train and test split is defined automatically. (20) Automated final model training on new unseen data (versioned features). The algorithm and hyperparameters are already predefined based on the settings of the previous experimentation stage. The model is retrained and refined. (21) Automated model evaluation and iterative adjustments of hyperparameters are executed, if required. Once the performance metrics indicate good results, the automated iterative training stops. The automated model training task and the automated model validation task can be iteratively repeated until a good result has been achieved. (22) The trained model is then exported and (23) pushed to the model registry (C6), where it is stored e.g., as code or containerized together with its associated configuration and environment files.

For all training job iterations, the ML metadata store (C7) records metadata such as parameters to train the model and the resulting performance metrics. This also includes the tracking and logging of the training job ID, training date and time, duration, and sources of artifacts. Additionally, the model specific metadata called "model lineage" combining the lineage of data and code is tracked for each newly registered model. This includes the source and version of the feature data and model training code used to train the model. Also, the model version and status (e.g., staging or production-ready) is recorded.

Once the status of a well-performing model is switched from staging to production, it is automatically handed over to the DevOps engineer or ML engineer for model deployment. From there, the (24) CI/CD component (C1) triggers the continuous deployment pipeline. The production-ready ML model and the model serving code are pulled (initially prepared by the software engineer (R5)). The continuous deployment pipeline carries out the build and test step of the ML model and serving code and deploys the model for production serving. The (25) model serving component (C8) makes predictions on new, unseen data coming from the feature store system (C4). This component can be designed by the software engineer (R5) as online inference for real-time predictions or as batch inference for predictions concerning large volumes of input data. For real-time predictions, features must come from the online database (low latency), whereas for batch predictions, features can be served from the offline database (normal latency). Model-serving applications are often configured within a container and prediction requests are handled via a REST API. As a foundational requirement, the ML engineer (R7) manages the model-serving computation infrastructure. The (26) monitoring component (C9) observes continuously the model-serving performance and infrastructure in real-time. Once a certain threshold is reached, such as detection of low prediction accuracy, the information is forwarded via the feedback loop. The (27) feedback loop is connected to the monitoring component (C9) and ensures fast and direct feedback allowing for more robust and improved predictions. It enables continuous training, retraining, and improvement. With the support of the feedback loop, information is transferred from the model monitoring component to several upstream receiver points, such as the experimental stage, data engineering zone, and the scheduler (trigger). The feedback to the experimental stage is taken forward by the data scientist for further model improvements. The feedback to the data engineering zone allows for the adjustment of the features prepared for the feature store system. Additionally, the detection of concept drifts as a feedback mechanism can enable (28) continuous training. For instance, once the model-monitoring component (C9) detects a drift in the data [3], the information is forwarded to the scheduler, which then triggers the automated ML workflow pipeline for retraining (continuous training). A change in adequacy of the deployed model can be detected using distribution comparisons to identify drift. Retraining is not only triggered automatically when a statistical threshold is reached; it can also be triggered when new feature data is available, or it can be scheduled periodically.

## 6   Conceptualization

With the findings at hand, we conceptualize the literature and interviews. It becomes obvious that the term MLOps is positioned at the intersection of machine learning, software engineering, DevOps, and data engineering (see Figure 5 in the Appendix). We define MLOps as follows:

*MLOps (Machine Learning Operations) is a paradigm, including aspects like best practices, sets of concepts, as well as a development culture when it comes to the end-to-end conceptualization, implementation, monitoring, deployment, and scalability of machine learning products. Most of all, it is an engineering practice that leverages three contributing disciplines: machine learning, software engineering (especially DevOps), and data engineering. MLOps is aimed at productionizing machine learning systems by bridging the gap between development (Dev) and operations (Ops). Essentially, MLOps aims to facilitate the creation of machine learning products by leveraging these principles: CI/CD automation, workflow orchestration, reproducibility; versioning of data, model, and code; collaboration; continuous ML training and evaluation; ML metadata tracking and logging; continuous monitoring; and feedback loops.*

## 7   Open Challenges

Several challenges for adopting MLOps have been identified after conducting the literature review, tool review, and interview study. These open challenges have been organized into the categories of organizational, ML system, and operational challenges.

**Organizational challenges.** The mindset and culture of data science practice is a typical challenge in organizational settings [2]. As our insights from literature and interviews show, to successfully develop and run ML products, there needs to be a culture shift away from model-driven machine learning toward a product-oriented discipline [γ]. The recent trend of data-centric AI also addresses this aspect by putting more focus on the data-related aspects taking place prior to the ML model building. Especially the roles associated with these activities should have a product-focused perspective when designing ML products [γ]. A great number of skills and individual roles are required for MLOps (β). As our identified sources point out, there is a lack of highly skilled experts for these roles—especially with regard to architects, data engineers, ML engineers, and DevOps engineers [29,41,44] [α, ε]. This is related to the necessary education of future professionals—as MLOps is typically not part of data science education [7] [γ]. Posoldova (2020) [35] further stresses this aspect by remarking that students should not only learn about model creation, but must also learn about technologies and components necessary to build functional ML products.

Data scientists alone cannot achieve the goals of MLOps. A multi-disciplinary team is required [14], thus MLOps needs to be a group process [α]. This is often hindered because teams work in silos rather than in cooperative setups [α]. Additionally, different knowledge levels and specialized terminologies make communication difficult. To lay the foundations for more fruitful setups, the respective decision-makers need to be convinced that an increased MLOps maturity and a product-focused mindset will yield clear business improvements [γ].

**ML system challenges.** A major challenge with regard to MLOps systems is designing for fluctuating demand, especially in relation to the process of ML training [7]. This stems from potentially voluminous and varying data [10], which makes it difficult to precisely estimate the necessary infrastructure resources (CPU, RAM, and GPU) and requires a high level of flexibility in terms of scalability of the infrastructure [7,26] [δ].

**Operational challenges**. In productive settings, it is challenging to operate ML manually due to different stacks of software and hardware components and their interplay. Therefore, robust automation is required [7,17]. Also, a constant incoming stream of new data forces retraining capabilities. This is a repetitive task which, again, requires a high level of automation [18] [θ]. These repetitive tasks yield a large number of artifacts that require a strong governance [24,29,40] as well as versioning of data, model, and code to ensure robustness and reproducibility [11,27,29]. Lastly, it is challenging to resolve a potential support request (e.g., by finding the root cause), as many parties and components are involved. Failures can be a combination of ML infrastructure and software [26].

## 8   Conclusion

With the increase of data availability and analytical capabilities, coupled with the constant pressure to innovate, more machine learning products than ever are being developed. However, only a small number of these proofs of concept progress into deployment and production. Furthermore, the academic space has focused intensively on machine learning model building and benchmarking, but too little on operating complex machine learning systems in real-world scenarios. In the real world, we observe data scientists still managing ML workflows manually to a great extent. The paradigm of Machine Learning Operations (MLOps) addresses these challenges. In this work, we shed more light on MLOps. By conducting a mixed-method study analyzing existing literature and tools, as well as interviewing eight experts from the field, we uncover four main aspects of MLOps: its principles, components, roles, and architecture. From these aspects, we infer a holistic definition. The results support a common understanding of the term MLOps and its associated concepts, and will hopefully assist researchers and professionals in setting up successful ML projects in the future.

# Appendix

**Table 1. List of evaluated technologies**

| | Technology Name | Description | Sources |
|---|---|---|---|
| **Open-source examples** | TensorFlow Extended | TensorFlow Extended (TFX) is a configuration framework providing libraries for each of the tasks of an end-to-end ML pipeline. Examples are data validation, data distribution checks, model training, and model serving. | [7,10,26,46] [δ, θ] |
| | Airflow | Airflow is a task and workflow orchestration tool, which can also be used for ML workflow orchestration. It is also used for orchestrating data engineering jobs. Tasks are executed according to directed acyclic graphs (DAGs). | [26,40,41] [α, β, ζ, η] |
| | Kubeflow | Kubeflow is a Kubernetes-based end-to-end ML platform. Each Kubeflow component is wrapped into a container and orchestrated by Kubernetes. Also, each task of an ML workflow pipeline is handled with one container. | [26,35,40,41,46] [α, β, γ, δ, ζ, η, θ] |
| | MLflow | MLflow is an ML platform that allows for the management of the ML lifecycle end-to-end. It provides an advanced experiment tracking functionality, a model registry, and model serving component. | [11,32,35] [α, γ, ε, ζ, η, θ] |
| **Commercial examples** | Databricks managed MLflow | The Databricks platform offers managed services based on other cloud providers' infrastructure, e.g., managed MLflow. | [26,32,35,40] [α, ζ] |
| | Amazon CodePipeline | Amazon CodePipeline is a CI/CD automation tool to facilitate the build, test, and delivery steps. It also allows one to schedule and manage the different stages of an ML pipeline. | [18] [γ] |
| | Amazon SageMaker | With SageMaker, Amazon AWS offers an end-to-end ML platform. It provides, out-of-the-box, a feature store, orchestration with SageMaker Pipelines, and model serving with SageMaker endpoints. | [7,11,18,24,35] [α, β, γ, ζ, θ] |
| | Azure DevOps Pipelines | Azure DevOps Pipelines is a CI/CD automation tool to facilitate the build, test, and delivery steps. It also allows one to schedule and manage the different stages of an ML pipeline. | [18,42] [γ, ε] |
| | Azure ML | Microsoft Azure offers, in combination with Azure DevOps Pipelines and Azure ML, an end-to-end ML platform. | [6,24,25,35,42] [α, γ, ε, ζ, η, θ] |

| | GCP - Vertex AI | GCP offers, along with Vertex AI, a fully managed end-to-end platform. In addition, they offer a managed Kubernetes cluster with Kubeflow as a service. | [25,35,40,41] [α, γ, δ, ζ, θ] |
|---|---|---|---|
| | IBM Cloud Pak for Data (IBM Watson Studio) | IBM Cloud Pak for Data combines a list of software in a package that offers data and ML capabilities. | [41] [γ] |

**Table 2. List of interview partners**

| Interviewee pseudonym | Job Title | Years of experience with DevOps | Years of experience with ML | Industry | Company Size (number of employees) |
|---|---|---|---|---|---|
| Alpha (α) | Senior Data Platform Engineer | 3 | 4 | Sporting Goods / Retail | 60,000 |
| Beta (β) | Solution architect / Specialist for ML and AI | 6 | 10 | IT Services / Cloud Provider / Cloud Computing | 25,000 |
| Gamma (γ) | AI Architect / Consultant | 5 | 7 | Cloud Provider | 350,000 |
| Delta (δ) | Technical Marketing & Manager in ML / AI | 10 | 5 | Cloud Provider | 139,995 |
| Epsilon (ε) | Technical Architect - Data & AI | 1 | 2 | Cloud Provider | 160,000 |
| Zeta (ζ) | ML engineering Consultant | 5 | 6 | Consulting Company | 569,000 |
| Eta (η) | Engineering Manager in AI / Senior Deep Learning Engineer | 10 | 10 | Conglomerate (multi-industry) | 400,000 |
| Theta (θ) | ML Platform Product Lead | 8 | 10 | Music / audio streaming | 6,500 |

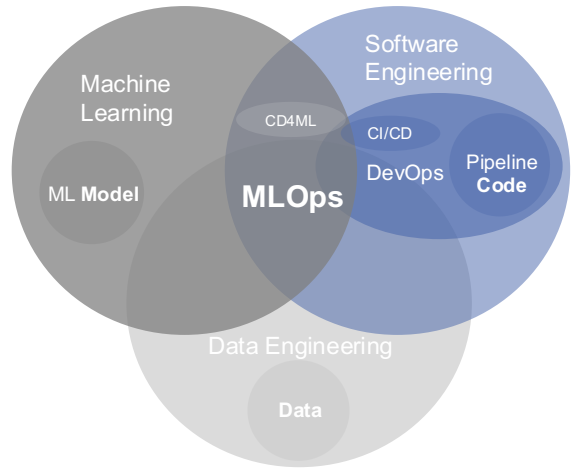

**Figure 5. Intersection of disciplines of the MLOps paradigm**